\documentclass{article}
\usepackage[preprint]{spconf} 
\usepackage{amsmath,graphicx,hyperref,multirow,amssymb,comment,bm}
\usepackage[table]{xcolor}
\definecolor{mygray}{gray}{.9}

\title{Shuffle PatchMix Augmentation with Confidence-Margin Weighted Pseudo-Labels for Enhanced Source-Free Domain Adaptation}
\name{Prasanna Reddy Pulakurthi$^{1}$, Majid Rabbani$^{1}$,
      Jamison Heard$^{1}$, Sohail Dianat$^{1}$,
      \thanks{This research was supported by DEVCOM Army Research Laboratory under contract W911QX-21-D-0001.}}
\namea{\textit{Celso M. de Melo}$^{2}$, and  \textit{Raghuveer Rao}$^{2}$}

\address{$^{1}$Rochester Institute of Technology, Rochester, NY, USA \\
$^{2}$DEVCOM Army Research Laboratory, Adelphi, MD, USA}

\copyrightnotice{%
\parbox[l]{2.0\columnwidth}{\footnotesize
\copyright\ 2025 IEEE. Personal use of this material is permitted. Permission from IEEE must be obtained for all other uses, in any current or future media, including reprinting/republishing this material for advertising or promotional purposes, creating new collective works, for resale or redistribution to servers or lists, or reuse of any copyrighted component of this work in other works.}}

\begin{document}
\maketitle

\begin{abstract}
This work investigates Source-Free Domain Adaptation (SFDA), where a model adapts to a target domain without access to source data. A new augmentation technique, Shuffle PatchMix (SPM), and a novel reweighting strategy are introduced to enhance performance. SPM shuffles and blends image patches to generate diverse and challenging augmentations, while the reweighting strategy prioritizes reliable pseudo-labels to mitigate label noise. These techniques are particularly effective on smaller datasets like PACS, where overfitting and pseudo-label noise pose greater risks. State-of-the-art results are achieved on three major benchmarks: PACS, VisDA-C, and DomainNet-126. Notably, on PACS, improvements of \textbf{7.3\%} (\textbf{79.4\%} to \textbf{86.7\%}) and \textbf{7.2\%} are observed in single-target and multi-target settings, respectively, while gains of \textbf{2.8\%} and \textbf{0.7\%} are attained on DomainNet-126 and VisDA-C. This combination of advanced augmentation and robust pseudo-label reweighting establishes a new benchmark for SFDA. The code is available at: \href{https://github.com/PrasannaPulakurthi/SPM}{\textbf{\textit{https://github.com/PrasannaPulakurthi/SPM}}}.

\end{abstract}


\vspace{-0.05in}
\section{Introduction}
\label{sec:intro}
\vspace{-0.05in}
Deep neural networks have achieved remarkable success in tasks where the training and test data share similar distributions. However, domain shifts can degrade network performance considerably \cite{ds1}. To address this challenge, Domain Adaptation (DA) techniques have been developed to improve model performance in the target domain by leveraging knowledge from a related source domain. The most studied DA paradigm, Unsupervised Domain Adaptation (UDA), adapts models from source to target domains where labeled target data is unavailable. A key strategy in UDA involves aligning feature distributions between source and target domains to reduce domain discrepancies \cite{adda, dann, cdan, famcd}. %

In many real-world scenarios, access to source data during adaptation is restricted due to privacy, security, or logistical constraints. This has led to Source-Free Domain Adaptation (SFDA), a subfield of UDA where a pre-trained source model is adapted to a target domain without source data. The lack of source data makes adaptation more challenging, requiring innovative techniques to align the model to the target domain while retaining knowledge learned from the source domain.

Advancements, such as AdaContrast \cite{adacontrast}, have demonstrated the effectiveness of self-training by pseudo-labeling, jointly trained with contrastive learning on strongly augmented target data to exploit the pairwise relationships among target samples. Despite their effectiveness, these methods heavily depend on the quality of pseudo-labels. If noisy pseudo-labels are treated equally, errors can be amplified, negatively impacting adaptation performance.

To address these limitations, this work proposes two key innovations: 
(1) \textbf{Confidence-Margin Reweighting} strategy that leverages both confidence (probability of the top-1 class prediction) and margin (difference between top-1 and top-2 class probabilities) of the pseudo-labels for self-training, which prioritizes reliable pseudo-labels to reduce the impact of label noise, and 
(2) \textbf{Shuffle PatchMix (SPM)} that is a novel augmentation strategy to enhance target domain representation by introducing diverse and challenging augmentations.
Together, these contributions lead to significant performance gains on three benchmark datasets.

\vspace{-0.05in}
\section{Related Work}
\label{sec:relatedwork}
\vspace{-0.05in}

\textbf{Domain Adaptation:} In UDA, several methods have been proposed to reduce domain discrepancies by aligning the feature distributions between source and target domains, MMD~\cite{mmd1}, MCC~\cite{mcc}, MDD~\cite{mdd}, and CMD~\cite{cmd}. Additionally, GAN-based methods use adversarial training to align the distributions in feature space \cite{adda,cdan} and image space \cite{cycada}. However, all these methods require access to the source data. In contrast, SFDA methods adapt to unlabeled target data without source samples. Several significant works have been proposed to address SFDA, including leveraging entropy minimization by TENT \cite{tent}, class prototypes by BAIT \cite{bait}, latent-prototype generation USFDA \cite{usfda}, and self-training by pseudo-labeling by SHOT \cite{shot}. Significant improvements are achieved by AdaContrast \cite{adacontrast} using self-training by pseudo-labels along with contrastive learning. 

\textbf{Learning with Noisy Pseudo-Labels:} The limitation of previous methods such as \cite{adacontrast} is that all the pseudo-labels equally contribute to the loss without considering the noise associated with the pseudo-labels. Various approaches were proposed to address noisy pseudo-labels. In semi-supervised learning, FixMatch \cite{fixmatch}, a fixed confidence threshold was used to filter pseudo-labels, while FreeMatch \cite{freematch} improved upon this by dynamically adapting the threshold based on prediction uncertainty. A meta-learning-based method \cite{ren} reweighted the loss but relied on a noise-free validation set, making it unsuitable for the SFDA setting. In SFDA, NEL \cite{nel} method employed a pseudo-label refinement framework that relied on an ensemble of pseudo-label predictions generated from different augmentations. While this method improved the robustness of pseudo-labels, it incurred a significant computational cost due to the need for multiple augmentations and predictions.
UPA~\cite{chen2024uncertainty} filters low-confidence pseudo-labels via uncertainty awareness, achieving the best results on VISDA-C.
In contrast, our reweighting strategy assigns continuous weights to pseudo-labels by leveraging both confidence and margin from aggregated neighboring predictions, ensuring more stable and reliable adaptation. 

\textbf{Data Augmentation:}
Data augmentation techniques are extensively utilized in domain adaptation to enhance model generalization, while contrastive learning aids in learning robust features. Recently, mixing-based data augmentation techniques have gained popularity, such as Mixup \cite{mixup}, CutMix \cite{cutmix}, Manifold-Mixup \cite{manifoldmixup}, Tokenmix \cite{tokenmix}, and TransMix \cite{transmix}. In CNN-based UDA tasks, several methods \cite{fixbi,pmtrans,adamixup} also use mixing-based strategies by linearly mixing the source and target domain data. 
In the SFDA setting, notable mixing-based methods such as ProxyMix \cite{proxymix} construct a proxy source domain by selecting confident target samples. Most recently, Improved SFDA~\cite{mitsuzumi2024understanding} introduced a learnable data augmentation via a teacher-student framework, whereas SF(DA)$^2$ \cite{sfda2} generates an augmentation graph in feature space to achieve the best results on DomainNet-126. 



Our proposed SPM augmentation differs from prior patch-shuffling methods~\cite{tokenmix, transmix} by seamlessly blending overlapping patches, effectively reducing blocking artifacts. In contrast to~\cite{tokenmix, transmix}, which mix patches across different images and were designed for supervised training, SPM operates intra-image and is inherently suited for source-free domain adaptation. By generating challenging adaptations, SPM outperforms previous methods to set a new state-of-the-art.


\vspace{-0.05in}
\section{Method}
\label{sec:method}
\vspace{-0.05in}
In the SFDA setting, the source model $g_s(.)$ is trained on labeled source data $\{\bm{x_s}^i,y_s^i\}_{i=1}^{n_s}$, where $\bm{x_s}^i \in \bm{\mathcal{X}_s}$ represent the source images and $y_s^i \in \bm{\mathcal{Y}_s}$ represent the source labels. During adaptation, the target images $\{\bm{x_t}^i\}_{i=1}^{n_t} \in \bm{\mathcal{X}_t}$ are available, while the underlying target labels $\{y_t^i\}_{i=1}^{n_t} \in \bm{\mathcal{Y}_t}$ are accessed only for evaluation. SFDA aims to train a model $g_t(.) = h_t(f_t(.))$, comprising of an encoder network $f_t(.)$ and a classification network $h_t(.)$, to classify the target data. During adaptation, the target model's parameters $\bm{\theta_t}$ are initialized with the source model's parameters $\bm{\theta_s}$.

\begin{figure}[tbp]
    \centering
    \includegraphics[width=0.99\columnwidth]{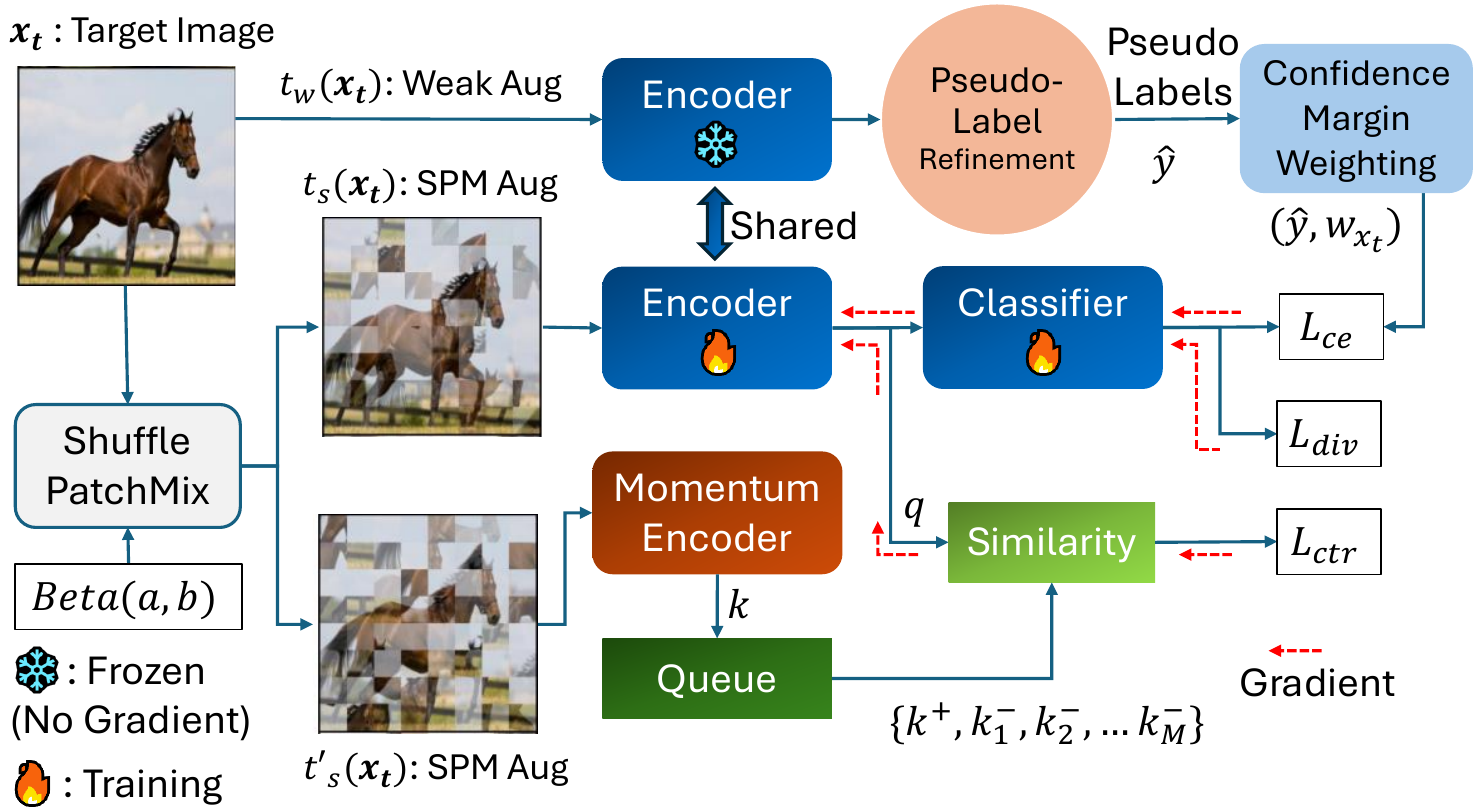}
    \vspace{-0.1in}
    \caption{An overview of the adaptation method.}
    \vspace{-0.1in}
    \label{fig:method}
\end{figure}

\vspace{0.05in}
\textbf{Adaptation Method Overview:} As illustrated in Figure \ref{fig:method}, the proposed adaptation method builds on the framework introduced in \cite{adacontrast}, with two key modifications: the integration of the SPM augmentation into the strong augmentation phase, and reweighting the self-training loss using pseudo-label confidence and margin estimation. The process begins by generating two strong augmentations, $t_s(\bm{x_t})$ and $t_s'(\bm{x_t})$, using SPM, along with a weak augmentation, $t_w(\bm{x_t})$. The weakly augmented image is processed through the encoder $f_t(.)$ to extract target features, which are then refined through the pseudo-labeling process to produce pseudo-labels, $\hat{y}_t$. The weight associated with each pseudo-label is determined using the proposed Confidence-Margin method. These refined pseudo-labels, $\hat{y}_t$, along with their weights, $w_{\bm{x_t}}$, are used to train the model $g_t(.)$ to classify the strongly augmented data, $t_s(\bm{x_t})$. The training optimizes a composite loss function that includes classification, diversity, and contrastive losses. This approach enables effective adaptation to the target domain by leveraging the diverse augmentations introduced by SPM and appropriately weighting the contribution of the pseudo-labels.

\begin{figure}[tbp]
    \centering
    \includegraphics[width=0.99\columnwidth]{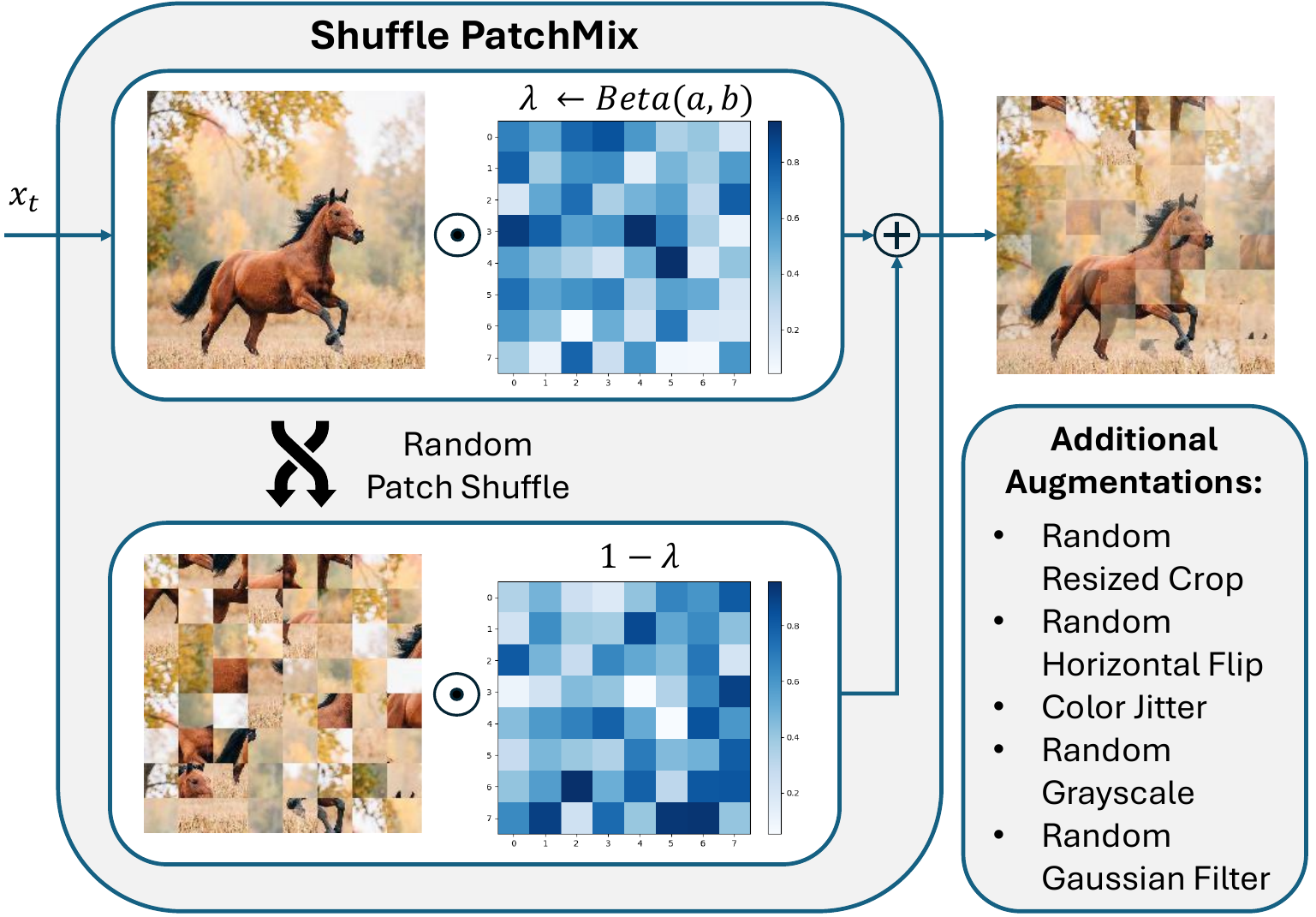}
    \vspace{-0.1in}
    \caption{Overview of the proposed Shuffle PatchMix method.} 
    \label{fig:spm}
\end{figure}

\vspace{0.05in}
\textbf{Shuffle PatchMix (SPM):} This work introduces SPM, a novel augmentation strategy that creates strong augmentations for contrastive learning. As illustrated in Figure \ref{fig:spm}, SPM first divides the target image into patches, denoted as $\bm{x_t^p}$. These patches are randomly shuffled to form $\bm{x_t^{sp}}$.
%
The original and shuffled patches are linearly combined using a scalar parameter $\lambda$, which is randomly generated from a Beta distribution. The Beta distribution has been successfully applied in the mixing-based augmentation method \cite{mixup,pmtrans}. Additional standard strong augmentations from \cite{moco_v2} are applied. 

%
The strength of patch mixing can be controlled with hyperparameters such as the number of patches $\nu$, 
and parameters of the Beta distribution $a$ and $b$. 
%
The mixing parameter $\lambda$ follows a Beta distribution, $\lambda \sim Beta(a,b)$, where $a,b > 0$, with the probability density function (PDF) given by:
\vspace{-0.05in}
\begin{equation}
f(\lambda; a, b) = \frac{\lambda^{a-1} (1-\lambda)^{b-1}}{\int_0^1 u^{a-1} (1-u)^{b-1} du} \quad \text{for } 0 \leq \lambda \leq 1.
\vspace{-0.05in}
\end{equation}
%
%

\begin{figure}[tb]
    \centering
    \includegraphics[width=0.99\columnwidth]{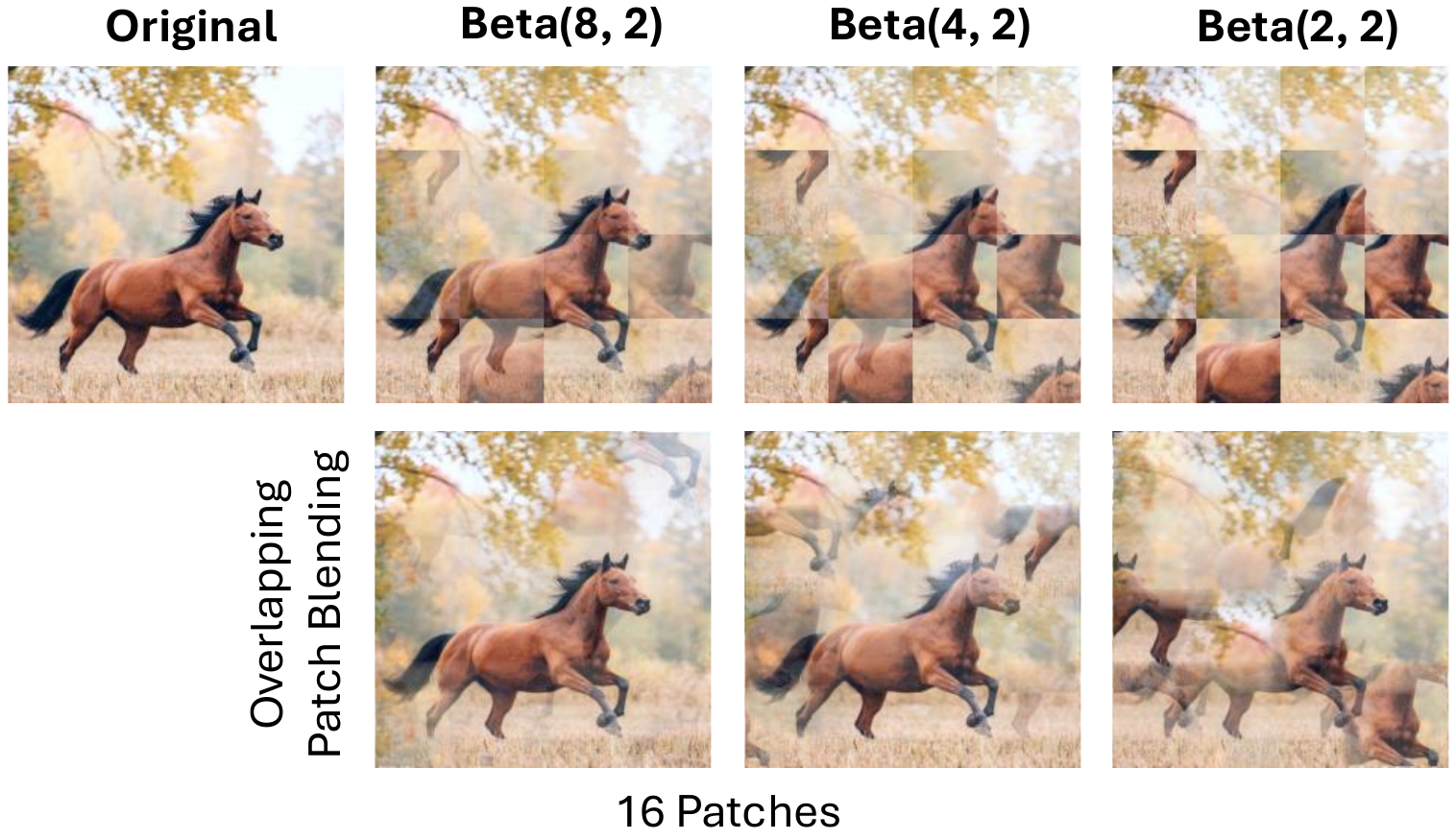}
    \vspace{-0.1in}
    \caption{(a) Different mixing strength SPM images (top). (b) SPM images with overlapping patch blending (bottom).} 
    \vspace{-0.05in}
    \label{fig:beta}
\end{figure}

The effects of varying patch mixing strengths, determined by different values of $a$ and $b$, are illustrated in Figure \ref{fig:beta}(a).

\textbf{Adaptive Mixing Strength:}
Our simulations demonstrate that gradually decreasing the mean of the random variable $\lambda$ during the training (i.e., increasing the mixing strength) allows the model to better adapt to the SPM augmentation distribution. To achieve this, the parameter $a$ in the Beta distribution $Beta(a,b)$ is progressively reduced from a start value $a_s$ to an end value $a_e$ ($a_e \leq a_s$) during the training.

\textbf{Mitigating Blocking Artifacts:}
As shown in Figure \ref{fig:beta}(a), patch shuffling introduces blocking artifacts. To address this, an overlapping patch blending approach is proposed, where patches are extracted with a linear dimension 30\% larger than the intended size and blended linearly with neighboring patches. This effectively reduces blocking artifacts, resulting in smoother transitions, as illustrated in Figure \ref{fig:beta}(b). Additional visualizations are provided in Appendix~\ref{sec:supp_spm_vis}.


Although applying SPM to all strongly augmented images can maximize diversity, two potential risks are introduced: reduced exposure to standard strong augmentations \cite{moco_v2}, and the generation of overly unrealistic samples that may hinder model training. Therefore, SPM is applied to a large fraction $\rho$ of the standard strong augmentations but not the entire set. 

\vspace{0.05in}
\textbf{Pseudo-label Refinement:} During adaptation, pseudo-labels are generated for the weakly augmented unlabeled target data $t_w(\bm{x_t})$ using the target model initialized with source weights, allowing knowledge transfer while gradually adapting to the target domain. This is performed on a per-batch basis through a nearest-neighbor soft voting strategy similar to \cite{adacontrast} using a memory queue $\bm{Q_w}$ representing the target feature space. For each target image, its weakly augmented version is encoded into a feature vector, which is then used to find the nearest neighbors in the target feature space. The pseudo-label is refined by averaging the probabilities of these neighbors, followed by an argmax operation.

To enable the nearest-neighbor search, the memory queue $\bm{Q_w}$ is maintained by storing features and probability predictions of the weakly augmented target samples, updated at each mini-batch. The feature space is stabilized using a slowly changing momentum model $g_t'(.)$. The momentum model's parameters $\bm{\theta_t'}$ are initialized with source weights $\bm{\theta_s}$, and are updated with momentum $m=0.999$ at each mini-batch step according to the update rule: $\bm{\theta_t'} \leftarrow m \bm{\theta_t'} + (1-m) \bm{\theta_t}$.

\vspace{0.05in}
\textbf{Loss Reweighting by Pseudo-Label Confidence-Margin Estimation:}
The pseudo-labels are obtained by averaging predictions of nearest neighbors and serve as self-supervised signals to classify strongly augmented images. Uniformly weighting all pseudo-labels can hinder adaptation, as noisy labels negatively affect the training process. 

To address this, we introduce the Confidence-Margin reweighting strategy, where \textbf{Confidence}, defined as the probability of the top prediction ($p_{\text{top1}}$), and \textbf{Margin} ($\Delta$), defined as the difference between the top two prediction probabilities ($\Delta = p_{\text{top1}} - p_{\text{top2}}$) for the pseudo-labels. If all the neighbors predominantly belong to the same class, the pseudo-label is considered highly reliable and should receive a high weight. This reliability is reflected in both the confidence and margin. A high margin corresponds to a more reliable pseudo-label, as it indicates a clear distinction between the most likely class and the second most likely class. Conversely, a low margin signals greater uncertainty and reduced reliability. 

The weights are computed as follows: 
\vspace{-0.05in}
\begin{equation}
\label{eq: 2}
    w_{\bm{x_t}} = \underbrace{p_{\text{top1}}}_{\text{confidence}}\;
        \underbrace{\vphantom{p_{\text{top1}}}\Delta}_{\text{margin}}\;
        \exp(\Delta).
\vspace{-0.05in}
\end{equation}
%

All symbols in Equation~\ref{eq: 2} are scalars, evaluated independently for each target image $\bm{x_t}$. The weighting function multiplies the confidence \(p_{\text{top1}}\) and the margin \(\Delta\), then scales their product by the exponential of margin \(\exp(\Delta)\).
The term $\exp({\Delta})$ supplies a smooth, margin-aware gain that exponentially enlarges the contribution of samples with large margins, whose pseudo-labels are statistically the most trustworthy. Samples with small margins are down-weighted, so that unreliable predictions exert little influence on the gradients. This continuous weighting eliminates the need for hand-tuned confidence thresholds and focuses learning on reliable targets, leading to faster and more stable adaptation.
Since pseudo-labels tend to be highly noisy in the early training stages, the reweighting strategy is gradually introduced as training progresses to ensure stability.


\vspace{0.05in}
\textbf{Weighted Classification Loss:} 
The refined pseudo-labels $\hat{y}_t$ generated from the weakly augmented data, along with their corresponding weights $w_{\bm{x_t}}$, are used to supervise the model's prediction for the strongly augmented version of the target data, as illustrated in Figure \ref{fig:method}. The proposed weighted classification loss is formulated as:
\vspace{-0.05in}
\begin{equation}
    L_{ce} = - \mathbb{E}_{\bm{x_t} \in \bm{\mathcal{X}_t}} 
    \left[w_{\bm{x_t}} \sum_{c=1}^{C} \hat{y_t}^c \log p_q^c\right],
\vspace{-0.05in}
\end{equation}
where $C$ is the number of classes and $p_q^c$ is the predicted probability of class $c$ for the strongly-augmented image $t_s(\bm{x_t})$.

\vspace{0.05in}
\textbf{Overall Loss:} The proposed weighted classification is jointly optimized with two additional losses from \cite{adacontrast}: the contrastive $L_{ctr}$ loss and the diversity $L_{div}$ loss. The total loss function is defined as:
\vspace{-0.05in}
\begin{equation}
\label{eq:5}
    L_t = L_{ce} + L_{ctr} + L_{div}.
\vspace{-0.05in}
\end{equation}
The diversity loss ($L_{div}$) serves as a regularization term to prevent model collapse by promoting diverse predictions. 
The contrastive loss ($L_{ctr}$) utilizes the SPM module to generate two augmentations, which are processed by the target and momentum encoders. A memory queue stores past keys, optimizing the model by pulling positive pairs closer while pushing negative pairs apart. Some of the images from the negative pairs may share the same pseudo-label; these same-class negative pairs are excluded. For more details refer to \cite{adacontrast}. 


\vspace{-0.05in}
\section{Experimental Setup}
\label{sec:expsetup}
\vspace{-0.05in}

\textbf{Datasets:}
The proposed method is evaluated on PACS \cite{pacs}, VisDA-C \cite{visda}, and DomainNet-126 \cite{domainnet}; results on Office-31~\cite{office31} and Office-Home~\cite{officehome} are given in Appendix~\ref{sec:supp_office}.
For DomainNet-126, seven domain shifts from four domains (Real, Sketch, Clipart, Painting) are constructed following \cite{adacontrast}, and the top-1 accuracy (\%) of each shift and their average are reported. For VisDA-C, the per-class top-1 accuracies (\%) and their average are compared.
For PACS, the single-target and multi-target protocols of \cite{nel} are followed.
Since \cite{adacontrast} did not report PACS results, its public code (verified on the results published in \cite{adacontrast}) was used to generate the baseline PACS results in Tables \ref{tab:pacs_single_target} and \ref{tab:pacs_multi_target}, as they outperformed the best-reported results in NEL \cite{nel}.

\textbf{Backbone:}
Following the standard DA protocols, all the experiments use ResNet-101 backbone for VisDA-C, ResNet-50 for DomainNet-126, and ResNet-18 for PACS. 

\vspace{0.05in}
\textbf{Hyperparameters:}
All hyperparameters are similar to those in \cite{adacontrast}, except for the following deviations, which were found to yield the optimal performance. The learning rate was fixed at $2\times10^{-4}$ for the SGD optimizer, the number of nearest neighbors was set to 3, and the model was trained for 100 epochs for PACS and 50 epochs for DomainNet-126 and VisDA-C. The starting value $a_s$ of the Beta distribution was set to 8 for DomainNet-126 and PACS, and 4 for VisDA-C, with a $\rho$ value of 0.8 used across all experiments. SPM randomly selected the number of patches $\nu$ from $\{2^2, 4^2, 8^2, 16^2\}$ in each mini-batch to enhance diversity.


\begin{table}[tb]
\vspace{-0.09in}
\caption{Classification accuracy (\%) on PACS for the single-target setting with ResNet-18. Legend: \textbf{P}: Photo, \textbf{A}: Art-Painting, \textbf{C}: Cartoon, and \textbf{S}: Sketch. The highest accuracies are in \textbf{bold}. * indicates our reproduced results using \cite{adacontrast}.}
\label{tab:pacs_single_target}
\vspace{-0.15in}
\begin{center}
\resizebox{\columnwidth}{!}{
\begin{tabular}{l|c c c c c c|c}
\hline

\hline
\rowcolor{mygray}
\textbf{Method} & \textbf{P→A} & \textbf{P→C} & \textbf{P→S} & \textbf{A→P} & \textbf{A→C} & \textbf{A→S} & \textbf{Avg.} \\
\hline
\hline
NEL \cite{nel} & 82.6 & 80.5 & 32.3 & 98.4 & 84.3 & 56.1 & 72.4 \\
AdaContrast \cite{adacontrast}*  & 81.3 & 72.2 & 66.7 & 98.7 & 79.7 & 77.9 & 79.4 \\
\hline
SPM (Ours) & \textbf{89.7} & \textbf{82.3} & \textbf{74.5} & \textbf{99.1} & \textbf{87.9} & \textbf{86.4} & \textbf{86.7} \\
\hline

\hline
\end{tabular}
}
\vspace{-0.25in}
\end{center}
\end{table}

\begin{table}[tb]
\caption{Classification accuracy (\%) on PACS for the multi-target setting with ResNet-18 backbone. The highest value is \textbf{bolded}. * indicates our reproduced results using \cite{adacontrast}.}
\label{tab:pacs_multi_target}
\vspace{-0.15in}
\begin{center}
\resizebox{\columnwidth}{!}{
\begin{tabular}{l|c|c c c|c c c|c}
\hline

\hline
\rowcolor{mygray}
\multicolumn{2}{c|}{\textbf{Multi-Target UDA}} & \multicolumn{3}{c|}{\textbf{P → A, C, S}} & \multicolumn{3}{c|}{\textbf{A → P, C, S}} & \\
\hline
\rowcolor{mygray}
\textbf{Method} & \textbf{SF} & \textbf{A} & \textbf{C} & \textbf{S} & \textbf{P} & \textbf{C} & \textbf{S} & \textbf{Avg.} \\
\hline
\hline
1-NN & $\boldsymbol{\times}$ & 15.2 & 18.1 & 25.6 & 22.7 & 19.7 & 22.7 & 20.7 \\
ADDA \cite{adda} & $\boldsymbol{\times}$ & 24.3 & 20.1 & 22.4 & 32.5 & 17.6 & 18.9 & 22.6 \\
DSN \cite{dsn} & $\boldsymbol{\times}$ & 28.4 & 21.1 & 25.6 & 29.5 & 25.8 & 26.8 & 25.8 \\
ITA \cite{ita} & $\boldsymbol{\times}$ & 31.4 & 23.0 & 28.2 & 35.7 & 27.0 & 28.9 & 29.0 \\
KD \cite{kd} & $\boldsymbol{\times}$ & 24.6 & 32.2 & 33.8 & 35.6 & 46.6 & 57.5 & 46.6 \\
NEL \cite{nel} & \checkmark & 80.1 & 76.1 & 25.9 & 96.0 & \textbf{82.8} & 49.8 & 68.4 \\
AdaContrast \cite{adacontrast}* & \checkmark & 70.1 & 77.9 & 62.9 & 95.9 & 72.7 & 72.9 & 75.4 \\
\hline
SPM (Ours) & \checkmark & \textbf{85.2} & \textbf{89.2} & \textbf{66.4} & \textbf{97.7} & 76.4 & \textbf{81.0} & \textbf{82.6} \\
\hline

\hline
\end{tabular}
}
\vspace{-0.25in}
\end{center}
\end{table}

\begin{table*}[tb]
\caption{Classification accuracy (\%) across various methods on VisDA-C (ResNet-101). SF stands for Source-Free. The highest values are shown in \textbf{bold} and the second highest are \underline{underlined}.} 
\label{tab:visda}
\vspace{-0.1in}
\begin{center}
\resizebox{\textwidth}{!}{
\begin{tabular}{l|c|c c c c c c c c c c c c|c}
\hline

\hline
\rowcolor{mygray}
\textbf{Method} & \textbf{SF} & \textbf{plane} & \textbf{bcycl} & \textbf{bus} & \textbf{car} & \textbf{horse} & \textbf{knife} & \textbf{mcycl} & \textbf{person} & \textbf{plant} & \textbf{sktbrd} & \textbf{train} & \textbf{truck} & \textbf{Avg.} \\
\hline
\hline
DANN \cite{dann} & $\boldsymbol{\times}$ & 81.9 & 77.7 & 82.8 & 44.3 & 81.2 & 29.5 & 65.1 & 28.6 & 51.9 & 54.6 & 82.8 & 7.8 & 57.4 \\
CDAN \cite{cdan} & $\boldsymbol{\times}$ & 85.2 & 66.9 & 83.0 & 50.8 & 84.2 & 74.9 & 88.1 & 74.5 & 83.4 & 76.0 & 81.9 & 38.0 & 73.9 \\
SWD \cite{swd} & $\boldsymbol{\times}$ & 90.8 & 82.5 & 81.7 & 70.5 & 91.7 & 69.5 & 86.3 & 77.5 & 87.4 & 63.6 & 85.6 & 29.2 & 76.4 \\
MCC \cite{mcc} & $\boldsymbol{\times}$ & 88.7 & 80.3 & 80.5 & 71.5 & 90.1 & 93.2 & 85.0 & 71.6 & 89.4 & 73.8 & 85.0 & 36.9 & 78.8 \\
CAN \cite{can} & $\boldsymbol{\times}$ & 97.0 & 87.2 & 82.5 & 74.3 & 97.8 & 96.2 & 90.8 & 80.7 & 96.6 & 96.3 & 87.5 & 59.9 & 87.2 \\
FixBi \cite{fixbi} & $\boldsymbol{\times}$ & 96.1 & 87.8 & 90.5 & 90.3 & 96.8 & 95.3 & 92.8 & 88.7 & 97.2 & 94.2 & 90.9 & 25.7 & 87.2 \\
\hline
Source only & - & 57.2 & 11.1 & 42.4 & 66.9 & 55.0 & 4.4 & 81.1 & 27.3 & 57.9 & 29.4 & 86.7 & 5.8 & 43.8 \\
MA \cite{ma} & \checkmark & 94.8 & 73.4 & 68.8 & 74.8 & 93.1 & 95.4 & 88.6 & 84.7 & 89.1 & 84.7 & 83.5 & 48.1 & 81.6 \\
BAIT \cite{bait} & \checkmark & 93.7 & 83.2 & 84.5 & 65.0 & 92.9 & 95.4 & 88.1 & 80.8 & 90.0 & 89.0 & 84.0 & 45.3 & 82.7 \\
SHOT \cite{shot} & \checkmark & 95.3 & 87.5 & 78.7 & 55.6 & 94.1 & 94.2 & 81.4 & 80.0 & 91.8 & 90.7 & 86.5 & 59.8 & 83.0 \\
AdaContrast \cite{adacontrast} & \checkmark & 97.0 & 84.7 & 84.0 & 77.3 & 96.7 & 93.8 & 91.9 & 84.8 & 94.3 & 93.1 & \underline{94.1} & 49.7 & 86.8 \\
SF(DA)$^2$ \cite{sfda2} & \checkmark & 96.8 & 89.3 & 82.9 & \underline{81.4} & 96.8 & 95.7 & 90.4 & 81.3 & 95.5 & 93.7 & 88.5 & \underline{64.7} & 88.1 \\
Improved SFDA \cite{mitsuzumi2024understanding} & \checkmark & \underline{97.5} & \textbf{91.4} & \textbf{87.9} & 79.4 & \underline{97.2} & \textbf{97.2} & \underline{92.2} & 83.0 & \underline{96.4} & 94.2 & 91.1 & 53.0 & 88.4 \\
UPA \cite{chen2024uncertainty} & \checkmark & 97.0 & \underline{90.4} & 82.6 & 65.0 & 96.7 & \underline{96.7} & 91.0 & \textbf{87.0} & \textbf{96.8} & \textbf{96.5} & 89.2 & \textbf{75.0} & \underline{88.7} \\
\hline
SPM (Ours) & \checkmark & \textbf{98.1} & 87.9 & \underline{86.7} & \textbf{86.2} & \textbf{97.7}  & 94.8 & \textbf{93.3}  & \underline{85.6} & 95.9 & \underline{95.6} & \textbf{95.5} & 55.3 & \textbf{89.4} \\
\hline

\hline
\end{tabular}
}
\vspace{-0.4in}
\end{center}
\end{table*}

\vspace{-0.05in}
\section{Results}
\label{sec:results}
\vspace{-0.05in}

\textbf{PACS Results:} 
Table \ref{tab:pacs_single_target} and Table \ref{tab:pacs_multi_target} summarize the performance of our proposed method on the PACS dataset under single-target and multi-target settings.
For the single-target setting, our method achieves an average accuracy of \textbf{86.7\%}, outperforming the baseline AdaContrast \cite{adacontrast} by a significant margin of \textbf{7.3\%}. The most notable improvements over baseline are observed in P→C (\textbf{82.3\%} vs. \textbf{72.2\%}) and P→S (\textbf{74.5\%} vs. \textbf{66.7\%}), demonstrating superior generalization across domain shifts.
In the multi-target setting, our method achieves the highest average accuracy of \textbf{82.6\%}, surpassing AdaContrast \cite{adacontrast} by \textbf{7.2\%}. Notable improvements are observed in challenging domain shifts, including P→A (\textbf{85.2\%} vs. \textbf{70.1\%}) and A→S (\textbf{81.0\%} vs. \textbf{72.9\%}), demonstrating the model's ability to adapt effectively to multiple target domains.

\vspace{0.05in}
\textbf{VisDA-C Results:} Table \ref{tab:visda} presents the classification accuracy of the VisDA-C dataset. Our method outperforms the best-performing method \cite{chen2024uncertainty} by \textbf{0.7\%}, achieving the highest average accuracy of \textbf{89.4\%}. Our method achieves the best or second-best performance in \textbf{8 out of 12} classes. 

\vspace{0.05in}
\textbf{DomainNet-126 Results:} Table \ref{tab:domainnet126} summarizes classification accuracy on the DomainNet-126 dataset. Our method achieves the highest average accuracy of \textbf{71.1\%} and outperforms the SOTA \cite{sfda2} by \textbf{2.8\%}. In addition, our method achieves the best performance on \textbf{5} of \textbf{7} domain shifts. 

\begin{table}[tb]
\caption{Classification accuracy (\%) on 7 domain shifts of the DomainNet-126 dataset (ResNet-50). Legend: \textbf{R}: Real, \textbf{C}: Clipart, \textbf{P}: Painting, and \textbf{S}: Sketch.}
\label{tab:domainnet126}
\vspace{-0.15in}
\begin{center}
\resizebox{\columnwidth}{!}{
\begin{tabular}{l|c|c c c c c c c|c}
\hline

\hline
\rowcolor{mygray}
\textbf{Method} & \textbf{SF} & \textbf{R→C} & \textbf{R→P} & \textbf{P→C} & \textbf{C→S} & \textbf{S→P} & \textbf{R→S} & \textbf{P→R} & \textbf{Avg.} \\
\hline
\hline
MCC \cite{mcc} & $\boldsymbol{\times}$ & 44.8 & 65.7 & 41.9 & 34.9 & 47.3 & 35.3 & 72.4 & 48.9 \\
\hline
Source only & - & 55.5 & 62.7 & 53.0 & 46.9 & 50.1 & 46.3 & 75.0 & 55.6 \\
TENT \cite{tent} & \checkmark & 58.5 & 65.7 & 57.9 & 48.5 & 52.4 & 54.0 & 67.0 & 57.7 \\
SHOT \cite{shot} & \checkmark & 67.7 & 68.4 & 66.9 & 60.1 & 66.1 & 59.9 & 80.8 & 67.1 \\
AdaContrast \cite{adacontrast} & \checkmark & \underline{70.2} & \underline{69.8} & \underline{68.6} & 58.0 & 65.9 & 61.5 & 80.5 & 67.8 \\
UPA \cite{chen2024uncertainty} & \checkmark & 68.6 & 69.5 & 67.6 & 60.9 & \underline{66.8} & 61.5 & \underline{80.9} & 68.0 \\
SF(DA)$^2$ \cite{sfda2} & \checkmark & 67.7 & 59.6 & 67.8 & \textbf{83.5} & 60.2 & \textbf{68.8} & 70.5 & \underline{68.3} \\
\hline
SPM (Ours) & \checkmark & \textbf{74.2} & \textbf{71.9} & \textbf{72.5} & \underline{62.4} & \textbf{68.1} & \underline{66.4} & \textbf{81.8} & \textbf{71.1} \\
\hline

\hline
\end{tabular}
}
\vspace{-0.3in}
\end{center}
\end{table}

\vspace{0.05in}
\textbf{Ablation Studies:} An ablation study was conducted on DomainNet-126 and PACS to assess the impact of SPM augmentation, patch overlap, and the reweighting strategy. The results, summarized in Table \ref{tab:ablation}, demonstrate the effectiveness of each component in improving classification accuracy.

The baseline AdaContrast achieves an accuracy of \textbf{67.8\%} on DomainNet-126 and \textbf{79.4\%} on PACS. When the reweighting strategy is applied, accuracy improves to \textbf{69.1\%} as unreliable pseudo-labels are down-weighted, while higher weights are assigned to more reliable ones during self-training. The SPM augmentation enhances performance by introducing challenging augmentations that encourage the network to learn class-invariant features, increasing accuracy to \textbf{70.2\%}. Additionally, patch overlapping mitigates blocking artifacts, resulting in a slight but consistent improvement to \textbf{70.4\%}.
Since the strength of SPM augmentation increases progressively during training, the model is trained for an extended period, as detailed in Section \ref{sec:expsetup}, allowing for better adaptation to varying augmentation distributions. When all components are combined, the highest accuracy of \textbf{71.1\%} on DomainNet-126 and \textbf{86.7\%} on PACS is achieved, demonstrating their complementary benefits in improving SFDA performance.

The largest improvement is observed in the smaller dataset, PACS, due to the following reasons. First, its limited size makes the model more prone to overfitting, so SPM is particularly effective in enhancing data diversity through patch mixing. Second, the impact of noisy pseudo‐labels is more pronounced in smaller datasets, making Confidence-Margin reweighting essential for reducing the influence of uncertain labels while emphasizing confident predictions.

\begin{table}[tb]
\caption{Ablation studies of sub-components of the proposed method measured by classification accuracy (\%) on DomainNet-126 and PACS.}
\label{tab:ablation}
\vspace{-0.2in}
\begin{center}
\resizebox{\columnwidth}{!}{
\begin{tabular}{c c c c |c |c}
\hline

\hline
\rowcolor{mygray}
\textbf{Baseline} & \textbf{SPM} & \textbf{Patch} & \textbf{Reweighting} & \textbf{DomainNet-126} & \textbf{PACS} \\
\rowcolor{mygray}
\cite{adacontrast} & \textbf{Augmentation} &  \textbf{Overlap} &  \textbf{Strategy} & \textbf{Avg.} & \textbf{Avg.} \\
\hline\checkmark & $\times$ & $\times$ & $\times$ & 67.8 & 79.4 \\
\checkmark & $\times$ & $\times$ & $ \checkmark$ & 69.1 & 81.8 \\
\checkmark & \checkmark & $\times$ & $\times$ & 70.2 & 83.8 \\
\checkmark & \checkmark  & \checkmark & $\times$ & 70.4 & 84.4\\
\checkmark & \checkmark  & \checkmark & \checkmark &  \textbf{71.1} & \textbf{86.7} \\
\hline

\hline
\end{tabular}
}
\vspace{-0.3in}
\end{center}
\end{table}

\vspace{-0.05in}
\section{Conclusion}
\label{sec:conclusion}
\vspace{-0.05in}
This paper introduces the SPM augmentation technique and the Confidence-Margin reweighting strategy to advance SFDA. Experimental results demonstrate that the proposed approach outperforms state-of-the-art methods across three benchmark datasets. Notably, the method is especially effective on smaller datasets like PACS, where limited target samples increase the risk of overfitting and label noise. The SPM augmentation technique enhances generalization by generating diverse training samples, while the reweighting strategy prioritizes reliable pseudo-labels, mitigating the impact of noisy predictions. Future work could explore extending SPM to broader domain adaptation paradigms, including self-supervised and semi-supervised learning.

\bibliographystyle{IEEEbib}
\footnotesize
\bibliography{main}

\clearpage
\normalsize
\appendix
%
\providecommand{\TBD}{--}  
\providecommand{\xmark}{$\boldsymbol{\times}$}

\providecommand{\maketitlesupplementary}{%
   \newpage
   \twocolumn[
    \centering
    \large
    \textbf{SHUFFLE PATCHMIX AUGMENTATION WITH CONFIDENCE-MARGIN WEIGHTED PSEUDO-LABELS FOR ENHANCED SOURCE-FREE DOMAIN ADAPTATION}\par
    \bigskip
    Supplementary Material\par
    \bigskip\bigskip
   ]%
}
\maketitlesupplementary

\section{Visualization of Shuffle PatchMix}
\label{sec:supp_spm_vis}

This section provides additional examples of SPM-augmented images, complementing Fig.~\ref{fig:beta} of the main paper. The mixing weight $\lambda \sim Beta(a,b)$ is drawn independently for each patch and weights the original patch, while $1-\lambda$ weights the shuffled patch. Unless stated otherwise, overlapping patch blending is used.

\begin{figure*}[t]
\centering
\begin{minipage}[t]{0.49\textwidth}
  \centering
  \includegraphics[width=\linewidth]{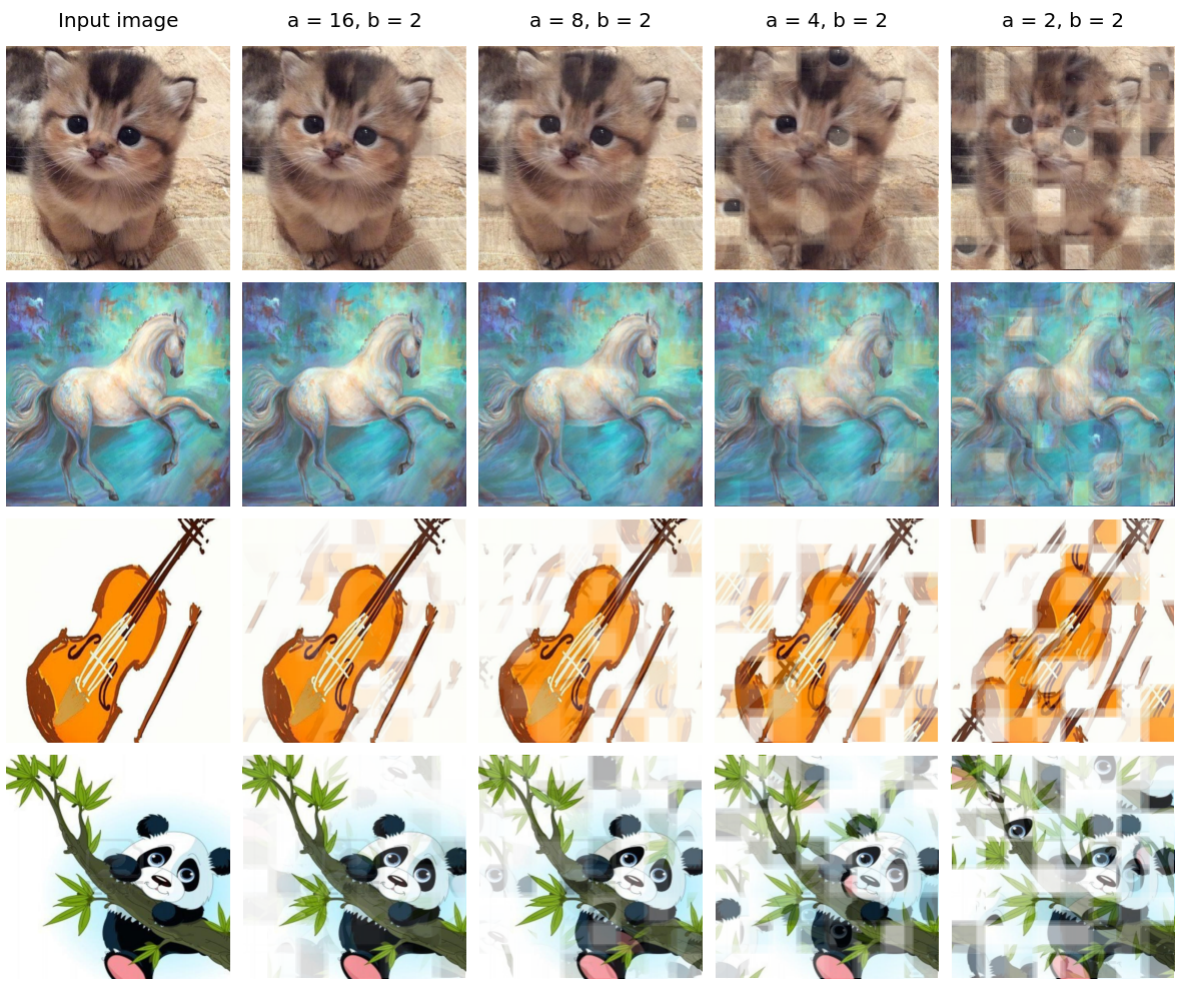}
  \caption{SPM images with different mixing strengths ($b=2$, $\nu=8^2$).}
  \label{fig:supp_spm_strength}
\end{minipage}\hfill
\begin{minipage}[t]{0.49\textwidth}
  \centering
  \includegraphics[width=\linewidth]{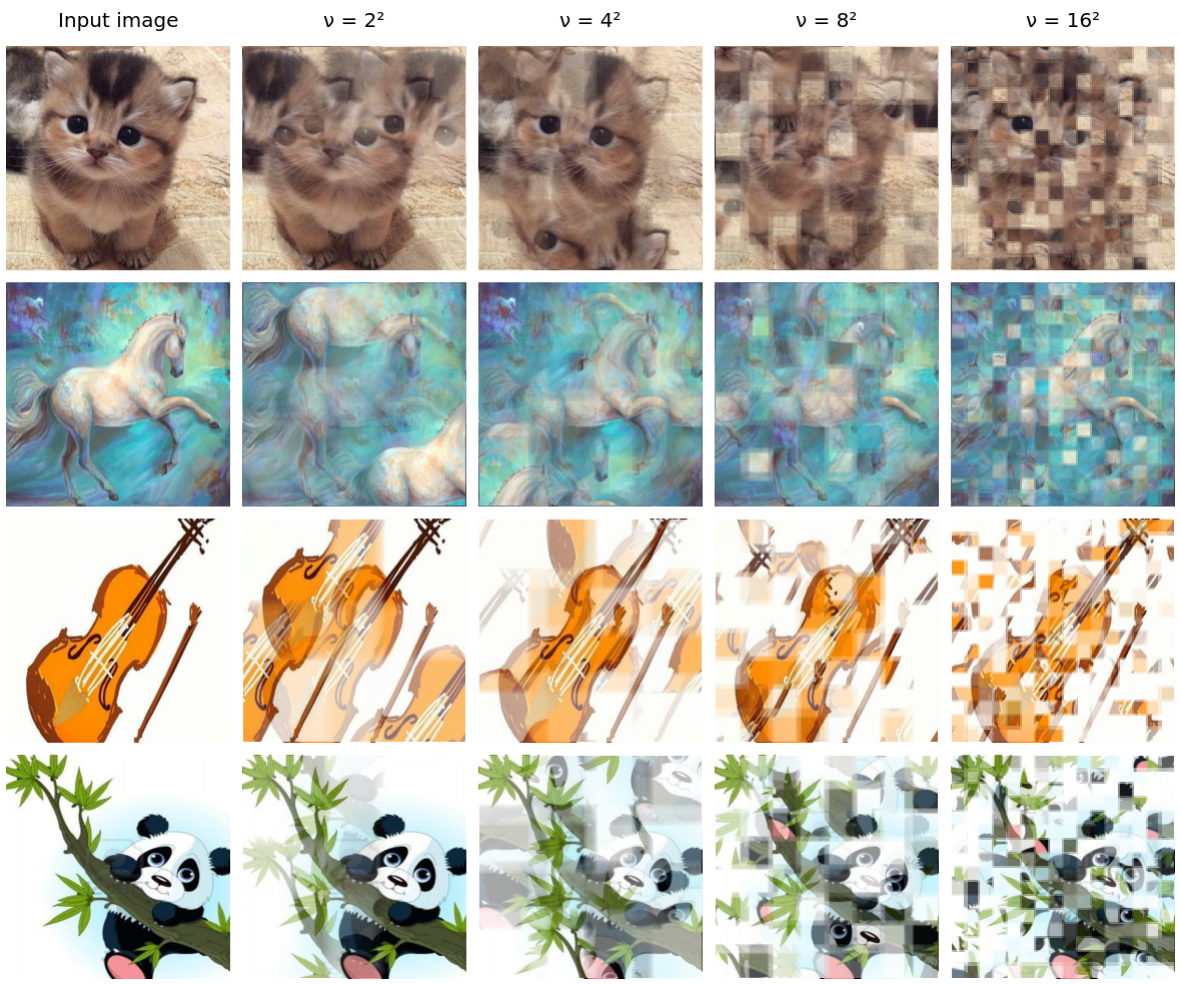}
  \caption{SPM images with different numbers of patches $\nu$ ($a=b=2$).}
  \label{fig:supp_spm_patches}
\end{minipage}
\end{figure*}

\textbf{Mixing strength:} Fig.~\ref{fig:supp_spm_strength} varies $a$ with $b=2$ and $\nu=8^2$ patches. Since the mean of $\lambda$ is $a/(a+b)$, a smaller $a$ increases the contribution of the shuffled patches. At $a=16$, the images are nearly unchanged, while at $a=2$ the shuffled patches are clearly visible and the object is only partially preserved. This motivates the schedule of the main paper, in which $a$ is decreased from $a_s$ to $a_e$ so that the augmentation becomes progressively stronger during adaptation.

\textbf{Number of patches:} Fig.~\ref{fig:supp_spm_patches} varies the number of patches $\nu$ with $a=b=2$. With few patches ($\nu=2^2$), large object parts are moved and duplicated, preserving part-level structure while changing the global layout. With many patches ($\nu=16^2$), the global shape is largely destroyed and mainly local texture and color remain. Sampling $\nu$ from $\{2^2, 4^2, 8^2, 16^2\}$ therefore exposes the model to augmentations at several spatial scales.

\begin{figure*}[t]
\centering
\includegraphics[width=0.9\textwidth]{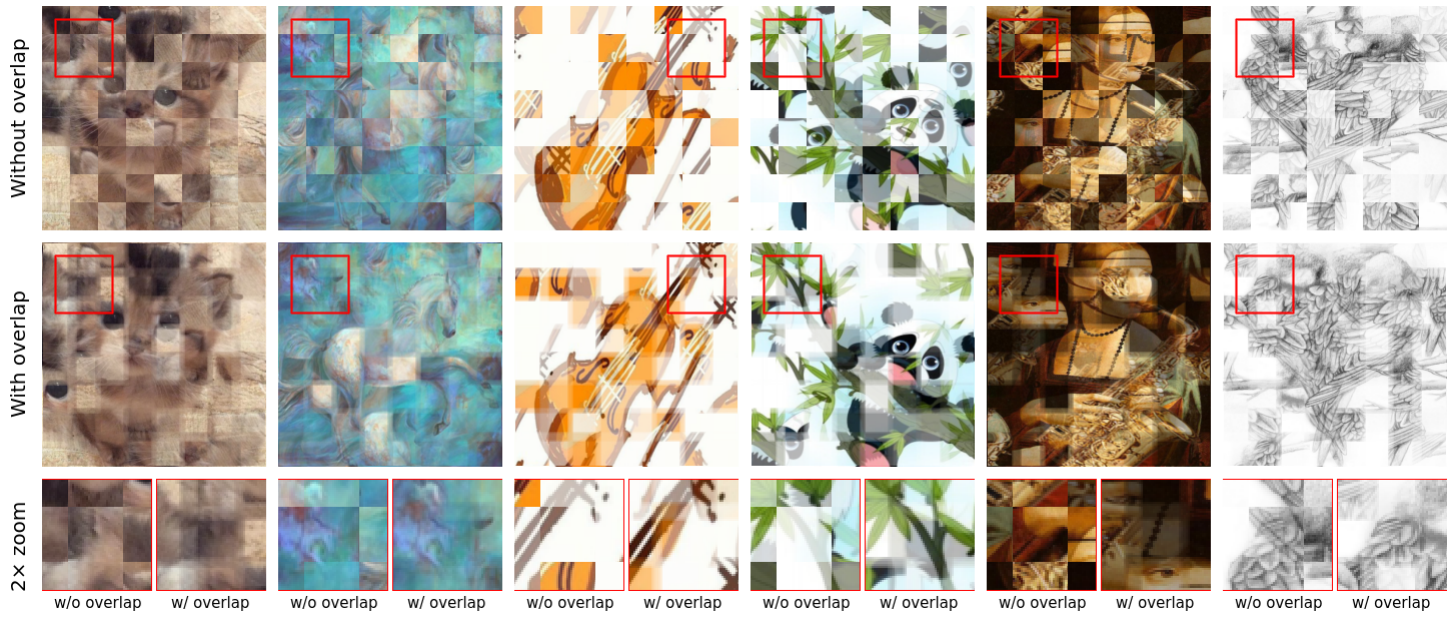}
\caption{SPM images without and with overlapping patch blending ($a=b=2$, $\nu=8^2$); the bottom row magnifies the region marked in red ($2\times$).}
\label{fig:supp_spm_overlap}
\end{figure*}

\textbf{Patch overlap:} Fig.~\ref{fig:supp_spm_overlap} compares non-overlapping and overlapping patch blending. Without overlap, the shuffled patches create sharp block boundaries that do not occur in natural images. Overlapping blending smooths these transitions while keeping the same degree of shuffling.

\section{Experiments on Office-31 and Office-Home}
\label{sec:supp_office}

To further assess the generality of SPM, the proposed method is evaluated on two additional standard SFDA benchmarks, Office-31 and Office-Home, which were not included in the main paper.

\subsection{Experimental Setup}
\label{sec:supp_office_setup}

\textbf{Datasets:}
Office-31~\cite{office31} contains 4,110 images of 31 object categories collected from three domains: Amazon (\textbf{A}, 2,817 images), DSLR (\textbf{D}, 498 images), and Webcam (\textbf{W}, 795 images), yielding six domain shifts.
Office-Home~\cite{officehome} contains 15,588 images of 65 categories from four domains: Art (\textbf{Ar}, 2,427), Clipart (\textbf{Cl}, 4,365), Product (\textbf{Pr}, 4,439), and Real-World (\textbf{Rw}, 4,357), yielding twelve domain shifts.
Both datasets are considerably smaller than VisDA-C and DomainNet-126, and Office-31 contains target domains with fewer than 500 images. They are therefore a natural setting for SPM, which was shown in the main paper to be most beneficial on small datasets such as PACS.

\textbf{Compared Methods:}
SPM is compared with a representative subset of the UDA and SFDA methods used in the main paper, all with a ResNet-50 backbone. UDA methods (\xmark{} in the SF column) use the labeled source data during adaptation, whereas SFDA methods (\checkmark) do not.
Results of the compared methods are taken from their original papers. When the original paper does not report a benchmark with ResNet-50, the ResNet-50 results reported in a later work are used instead: on Office-31, DANN from~\cite{cdan}, SWD from~\cite{daln}, and AdaContrast from~\cite{difo}; on Office-Home, DANN from~\cite{cdan}, MCC from~\cite{scda}, CAN from the supplementary material of~\cite{srdc}, AdaContrast from~\cite{difo}, and SF(DA)$^2$ from~\cite{silan}. CDAN denotes the CDAN+E variant~\cite{cdan}.

\textbf{Implementation Details:}
Following the standard protocol for both benchmarks, a ResNet-50 backbone with a 256-dimensional bottleneck is used, identical to the DomainNet-126 setting.
For adaptation, the learning rate of $2\times10^{-4}$, 3 nearest neighbors, $a_s = 8$, $\rho = 0.8$, and $\nu \in \{2^2, 4^2, 8^2, 16^2\}$ follow the DomainNet-126 setting of the main paper, together with a batch size of 128, $b = 2$, and a minimum of 3,000 adaptation iterations. Office-Home uses these settings, except that adaptation runs for a fixed 88 epochs on all shifts (about 3,000 iterations on the larger target domains). On the much smaller Office-31, adaptation runs for 100 epochs, as for PACS in the main paper, with a batch size of 32. Because of the 3,000-iteration minimum, the DSLR and Webcam targets run for 200 and 125 epochs, respectively. The reported accuracy is the target accuracy at the end of adaptation.

\subsection{Results}
\label{sec:supp_office_results}

Tables~\ref{tab:office31} and~\ref{tab:officehome} report the results on Office-31 and Office-Home, respectively.
On Office-31, SPM achieves the best source-free average accuracy of 90.0\%, a gain of 10.5\% over the source-only model (79.5\%), ahead of the strongest source-free competitors SF(DA)$^2$~\cite{sfda2} and UPA~\cite{chen2024uncertainty} (89.9\%). SPM obtains the best source-free result on D→A (77.7\%, +2.0\% over the next best) and ties for the best on D→W.
On Office-Home, SPM achieves an average accuracy of 71.7\%, improving over the source-only model (60.3\%) by 11.4\%, but remaining below UPA~\cite{chen2024uncertainty} (73.0\%).
The largest gains over the source model are on A→W (+18.3\%) and D→A (+16.1\%) on Office-31, and on Cl→Ar (+15.9\%) and Cl→Pr (+15.1\%) on Office-Home.
SPM is considerably closer to the state of the art on the smaller Office-31 than on Office-Home, consistent with the observation in the main paper that SPM is most beneficial on small target domains.

\begin{table}[tb]
\caption{Accuracy (\%) on Office-31 (ResNet-50). \textbf{A}: Amazon, \textbf{D}: DSLR, \textbf{W}: Webcam; SF: source-free.}
\label{tab:office31}
\begin{center}
\resizebox{\columnwidth}{!}{
\begin{tabular}{l|c|c c c c c c|c}
\hline

\hline
\rowcolor{mygray}
\textbf{Method} & \textbf{SF} & \textbf{A→D} & \textbf{A→W} & \textbf{D→A} & \textbf{D→W} & \textbf{W→A} & \textbf{W→D} & \textbf{Avg.} \\
\hline
\hline
DANN \cite{dann} & \xmark & 79.7 & 82.0 & 68.2 & 96.9 & 67.4 & 99.1 & 82.2 \\
CDAN \cite{cdan} & \xmark & 92.9 & 94.1 & 71.0 & 98.6 & 69.3 & 100.0 & 87.7 \\
SWD \cite{swd} & \xmark & 94.7 & 90.4 & 70.3 & 98.7 & 70.5 & 100.0 & 87.4 \\
MCC \cite{mcc} & \xmark & 94.4 & 95.5 & 72.9 & 98.6 & 74.9 & 100.0 & 89.4 \\
CAN \cite{can} & \xmark & 95.0 & 94.5 & 78.0 & 99.1 & 77.0 & 99.8 & 90.6 \\
FixBi \cite{fixbi} & \xmark & 95.0 & 96.1 & 78.7 & 99.3 & 79.4 & 100.0 & 91.4 \\
\hline
Source only & - & 81.5 & 75.7 & 61.6 & 96.0 & 63.2 & 99.0 & 79.5 \\
MA \cite{ma} & \checkmark & 92.7 & 93.7 & 75.3 & \underline{98.5} & \textbf{77.8} & 99.8 & 89.6 \\
BAIT \cite{bait} & \checkmark & 92.0 & \textbf{94.6} & 74.6 & 98.1 & 75.2 & \textbf{100.0} & 89.1 \\
SHOT \cite{shot} & \checkmark & 94.0 & 90.1 & 74.7 & 98.4 & 74.3 & \underline{99.9} & 88.6 \\
AdaContrast \cite{adacontrast} & \checkmark & 87.7 & 83.1 & 73.7 & 91.3 & \underline{77.6} & 72.8 & 81.0 \\
SF(DA)$^2$ \cite{sfda2} & \checkmark & \underline{95.8} & 92.1 & \underline{75.7} & \textbf{99.0} & 76.8 & 99.8 & \underline{89.9} \\
UPA \cite{chen2024uncertainty} & \checkmark & \textbf{96.2} & 93.7 & 75.0 & 98.4 & 76.6 & 99.8 & \underline{89.9} \\
\hline
SPM (Ours) & \checkmark & \underline{95.8} & \underline{94.0} & \textbf{77.7} & \textbf{99.0} & 74.5 & 99.4 & \textbf{90.0} \\
\hline

\hline
\end{tabular}
}
\end{center}
\end{table}

\begin{table*}[tb]
\caption{Accuracy (\%) on Office-Home (ResNet-50). \textbf{Ar}: Art, \textbf{Cl}: Clipart, \textbf{Pr}: Product, \textbf{Rw}: Real-World; SF: source-free.}
\label{tab:officehome}
\begin{center}
\resizebox{\textwidth}{!}{
\begin{tabular}{l|c|c c c c c c c c c c c c|c}
\hline

\hline
\rowcolor{mygray}
\textbf{Method} & \textbf{SF} & \textbf{Ar→Cl} & \textbf{Ar→Pr} & \textbf{Ar→Rw} & \textbf{Cl→Ar} & \textbf{Cl→Pr} & \textbf{Cl→Rw} & \textbf{Pr→Ar} & \textbf{Pr→Cl} & \textbf{Pr→Rw} & \textbf{Rw→Ar} & \textbf{Rw→Cl} & \textbf{Rw→Pr} & \textbf{Avg.} \\
\hline
\hline
DANN \cite{dann} & \xmark & 45.6 & 59.3 & 70.1 & 47.0 & 58.5 & 60.9 & 46.1 & 43.7 & 68.5 & 63.2 & 51.8 & 76.8 & 57.6 \\
CDAN \cite{cdan} & \xmark & 50.7 & 70.6 & 76.0 & 57.6 & 70.0 & 70.0 & 57.4 & 50.9 & 77.3 & 70.9 & 56.7 & 81.6 & 65.8 \\
MCC \cite{mcc} & \xmark & 55.1 & 75.2 & 79.5 & 63.3 & 73.2 & 75.8 & 66.1 & 52.1 & 76.9 & 73.8 & 58.4 & 83.6 & 69.4 \\
CAN \cite{can} & \xmark & 58.5 & 75.3 & 75.1 & 61.7 & 74.5 & 70.1 & 61.3 & 54.6 & 75.9 & 72.4 & 58.3 & 82.4 & 68.3 \\
FixBi \cite{fixbi} & \xmark & 58.1 & 77.3 & 80.4 & 67.7 & 79.5 & 78.1 & 65.8 & 57.9 & 81.7 & 76.4 & 62.9 & 86.7 & 72.7 \\
\hline
Source only & - & 44.1 & 67.2 & 74.4 & 52.3 & 63.1 & 64.6 & 53.4 & 41.1 & 73.7 & 65.5 & 46.8 & 77.9 & 60.3 \\
BAIT \cite{bait} & \checkmark & 57.4 & 77.5 & \textbf{82.4} & 68.0 & 77.2 & 75.1 & 67.1 & \underline{55.5} & 81.9 & \underline{73.9} & 59.5 & 84.2 & 71.6 \\
SHOT \cite{shot} & \checkmark & 57.1 & 78.1 & \underline{81.5} & 68.0 & 78.2 & 78.1 & \underline{67.4} & 54.9 & \underline{82.2} & 73.3 & 58.8 & 84.3 & 71.8 \\
AdaContrast \cite{adacontrast} & \checkmark & 47.2 & 75.1 & 75.5 & 60.7 & 73.3 & 73.2 & 60.2 & 45.2 & 76.6 & 65.6 & 48.3 & 79.1 & 65.0 \\
SF(DA)$^2$ \cite{sfda2} & \checkmark & \underline{57.8} & \textbf{80.2} & \underline{81.5} & \textbf{69.5} & \underline{79.2} & \underline{79.4} & 66.5 & \textbf{57.2} & 82.1 & 73.3 & \textbf{60.2} & 83.8 & \underline{72.6} \\
UPA \cite{chen2024uncertainty} & \checkmark & 57.3 & \underline{79.5} & \textbf{82.4} & \underline{68.2} & \textbf{81.0} & \textbf{80.2} & \textbf{69.3} & 55.4 & \textbf{82.3} & \textbf{75.3} & 59.2 & \textbf{85.5} & \textbf{73.0} \\
\hline
SPM (Ours) & \checkmark & \textbf{58.2} & 78.8 & 79.6 & \underline{68.2} & 78.2 & 78.5 & \underline{67.4} & 55.1 & 81.6 & 70.3 & \underline{59.9} & \underline{85.0} & 71.7 \\
\hline

\hline
\end{tabular}
}
\end{center}
\end{table*}

\section{Ablation of the Pseudo-Label Weighting Function}
\label{sec:supp_reweighting}

Table~\ref{tab:supp_reweighting} compares different weighting functions $w_{\bm{x_t}}$ for the pseudo-label loss on PACS in the single-target setting. All other components are identical to the full method, including SPM augmentation with patch overlap, so only the weight assigned to each pseudo-label changes. Uniform weights ($w_{\bm{x_t}}=1$) correspond to the SPM and patch overlap configuration in the ablation of the main paper (84.4\%).

Weighting by either the margin $\Delta$ or the confidence $p_{\text{top1}}$ alone already improves the average accuracy by 1.3\% and 1.4\%, respectively, confirming that down-weighting unreliable pseudo-labels is beneficial. The proposed confidence-margin weight $p_{\text{top1}}\,\Delta\,\exp(\Delta)$ (Equation~\ref{eq: 2} of the main paper) achieves the best average accuracy of 86.7\%, a gain of 2.3\% over uniform weights, and is the best or tied for the best on five of the six shifts.

\begin{table}[tb]
\caption{Accuracy (\%) on PACS (single-target, ResNet-18) with different pseudo-label weighting functions. \textbf{P}: Photo, \textbf{A}: Art Painting, \textbf{C}: Cartoon, \textbf{S}: Sketch.}
\label{tab:supp_reweighting}
\begin{center}
\resizebox{\columnwidth}{!}{
\begin{tabular}{l|c c c c c c|c}
\hline

\hline
\rowcolor{mygray}
\textbf{Weight} $w_{\bm{x_t}}$ & \textbf{P→A} & \textbf{P→C} & \textbf{P→S} & \textbf{A→P} & \textbf{A→C} & \textbf{A→S} & \textbf{Avg.} \\
\hline
\hline
$1$ (uniform) & 86.6 & 77.7 & \underline{74.1} & \textbf{99.1} & 84.1 & \underline{84.6} & 84.4 \\
$\Delta$ (margin) & \underline{89.5} & \textbf{82.7} & 73.4 & \textbf{99.1} & \underline{86.4} & 83.1 & 85.7 \\
$p_{\text{top1}}$ (confidence) & 89.4 & 81.7 & 74.0 & \textbf{99.1} & \underline{86.4} & 84.3 & \underline{85.8} \\
$p_{\text{top1}}\,\Delta\,\exp(\Delta)$ (ours) & \textbf{89.7} & \underline{82.3} & \textbf{74.5} & \textbf{99.1} & \textbf{87.9} & \textbf{86.4} & \textbf{86.7} \\
\hline

\hline
\end{tabular}
}
\end{center}
\end{table}

\section{Grad-CAM Visualization}
\label{sec:supp_gradcam}

\begin{figure*}[t]
\centering
\includegraphics[width=\textwidth]{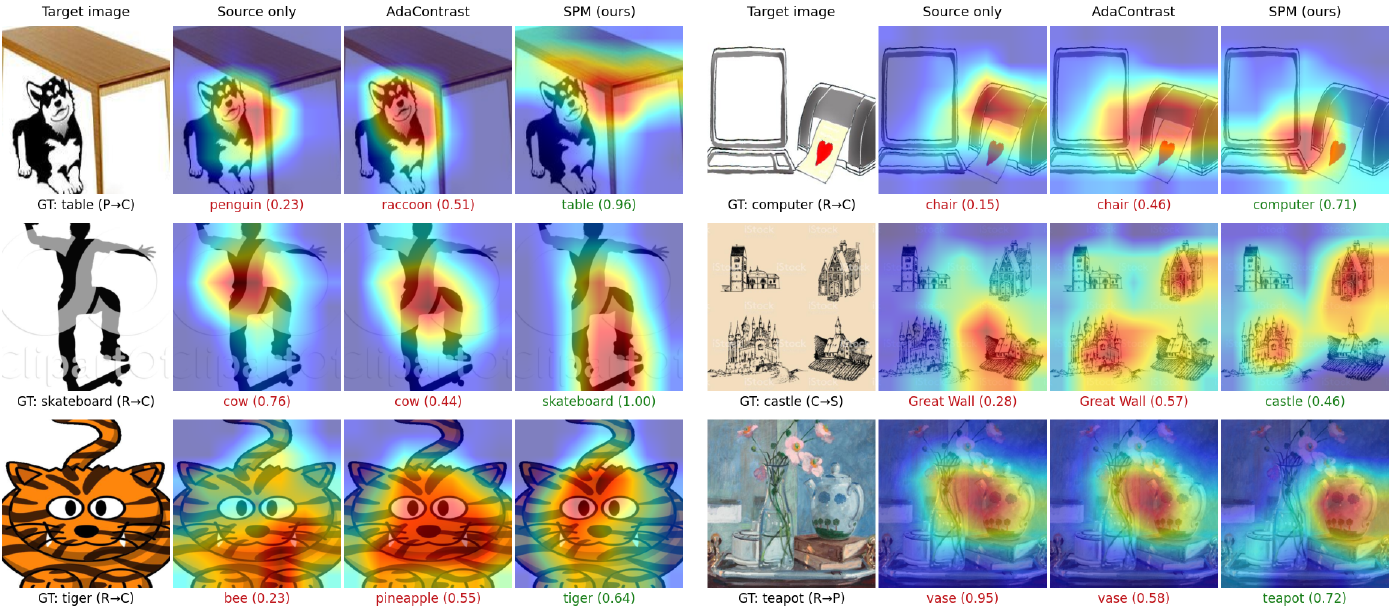}
\caption{Grad-CAM~\cite{gradcam} of the source-only model, AdaContrast, and SPM on DomainNet-126 for target images that only SPM classifies correctly (\textbf{P}: Painting, \textbf{R}: Real, \textbf{C}: Clipart, \textbf{S}: Sketch). Below each map: predicted class and confidence (green: correct, red: wrong).}
\label{fig:supp_gradcam}
\end{figure*}

\begin{figure*}[t]
\centering
\includegraphics[width=\textwidth]{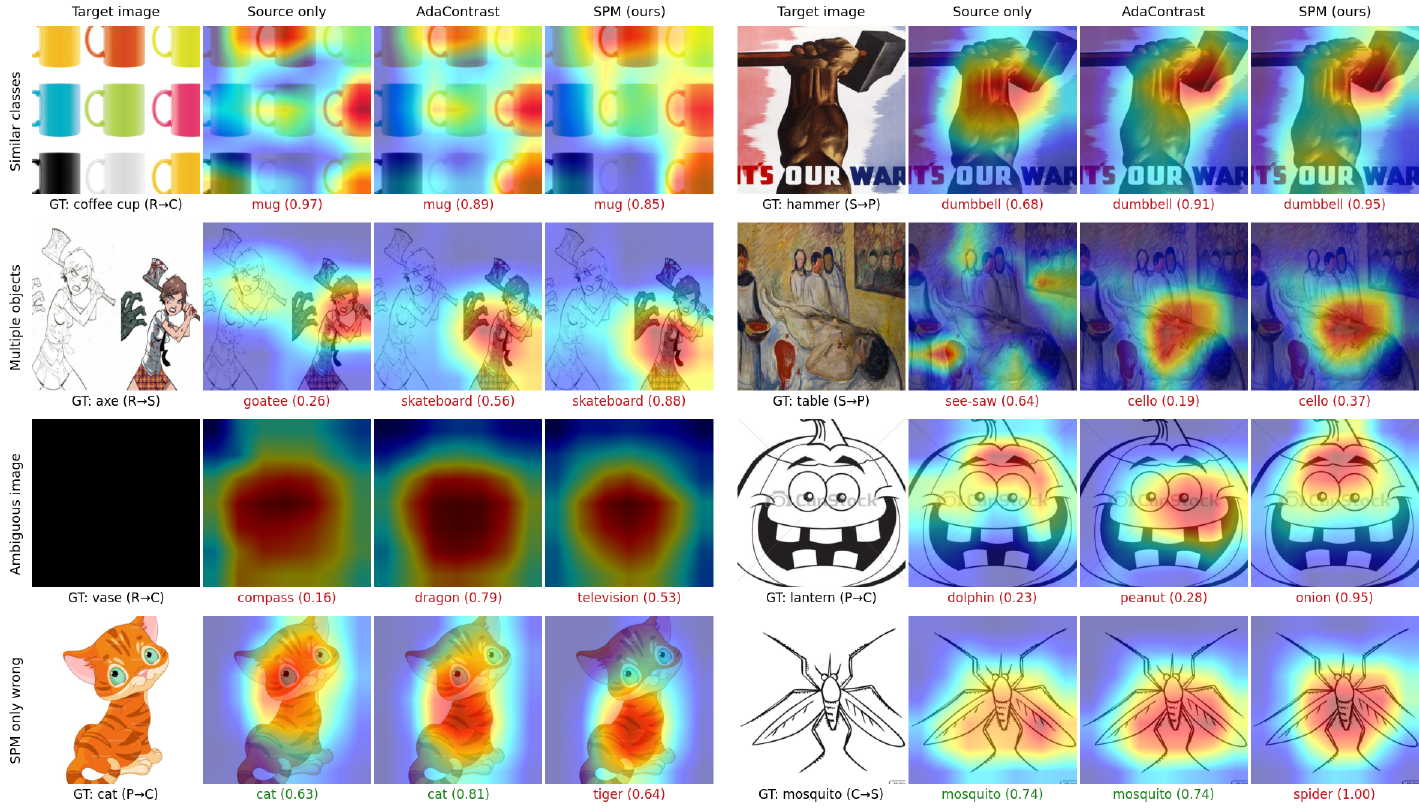}
\caption{Failure cases: Grad-CAM of the source-only model, AdaContrast, and SPM on DomainNet-126 target images that all three models misclassify, grouped by the apparent cause (first three rows), and on images that only SPM misclassifies (last row). \textbf{P}: Painting, \textbf{R}: Real, \textbf{C}: Clipart, \textbf{S}: Sketch. Below each map: predicted class and confidence (green: correct, red: wrong).}
\label{fig:supp_gradcam_fail}
\end{figure*}

Figs.~\ref{fig:supp_gradcam} and~\ref{fig:supp_gradcam_fail} visualize the image regions that drive the predictions of the source-only model, AdaContrast~\cite{adacontrast}, and SPM on DomainNet-126, using Grad-CAM~\cite{gradcam} on the last ResNet-50 block for each model's predicted class. The models are those evaluated in the main paper, and the examples are selected from random samples of the target images of each group.

\textbf{Improvements:} Fig.~\ref{fig:supp_gradcam} shows target images that only SPM classifies correctly. In these examples, the source-only model and AdaContrast attend to a salient but unrelated object or to a single part, e.g., the dog instead of the table, the printer instead of the computer, the rider instead of the skateboard, or the vase instead of the teapot, whereas SPM attends to the labeled object. This is consistent with the design of SPM: training on images with shuffled and blended patches discourages reliance on a single part or on the global layout of the scene.

\textbf{Failure cases:} Fig.~\ref{fig:supp_gradcam_fail} shows target images that all three models misclassify, grouped by the apparent cause. (i) \emph{Similar classes}: the prediction is a visually or semantically close class, such as \emph{mug} for a coffee cup or \emph{dumbbell} for a hammer. (ii) \emph{Multiple objects}: the image contains several objects and the labeled one is not the most salient, such as an axe held by a person or a table in a crowded scene. (iii) \emph{Ambiguous images}: the image is uninformative or its label is debatable, such as an all-black image labeled \emph{vase} or a jack-o'-lantern face labeled \emph{lantern}. In most of these cases, the three models attend to similar regions, indicating that the errors stem from class overlap, scene composition, or label noise in DomainNet-126 rather than from the adaptation itself. The last row shows the opposite case, images that the source-only model and AdaContrast classify correctly but SPM does not. These errors are usually confident confusions between classes that share local parts or texture, such as \emph{tiger} for a striped cat or \emph{spider} for a mosquito with spread legs, where SPM attends to the stripes or the legs. Because SPM is trained on shuffled patches, it relies more on local evidence and less on global shape, which helps in most cases (Fig.~\ref{fig:supp_gradcam}) but can fail when the classes differ mainly in their global shape.

\end{document}